# Bayesian Inference with Anchored Ensembles of Neural Networks, and Application to Exploration in Reinforcement Learning

Tim Pearce [1] [2]   Nicolas Anastassacos [2] [3]   Mohamed Zaki [1]   Andy Neely [1]

## Abstract

The use of ensembles of neural networks (NNs) for the quantification of predictive uncertainty is widespread. However, the current justification is intuitive rather than analytical. This work proposes one minor modification to the normal ensembling methodology, which we prove allows the ensemble to perform Bayesian inference, hence converging to the corresponding Gaussian Process as both the total number of NNs, and the size of each, tend to infinity.

This working paper provides early-stage results in a reinforcement learning setting, analysing the practicality of the technique for an ensemble of small, finite number. Using the uncertainty estimates produced by anchored ensembles to govern the exploration-exploitation process results in steadier, more stable learning.

## 1. Introduction

### 1.1. Uncertainty and Neural Networks

By many measures neural networks (NNs) have become the dominant approach within machine learning, having achieved state-of-the-art results across domains. However, they are not probabilistic in nature, which makes understanding the certainty (or uncertainty) of individual predictions a challenge.

This is important because uncertainty quantification is vital for many real-world applications. A principled approach to dealing with uncertainty is provided by the Bayesian framework (Ghahramani, 2015), which allows uncertainty in parameters to be modelled, and hence predictive uncertainty expressed. Bayesian Neural Networks (BNNs) are NNs over which full Bayesian inference is performed (MacKay, 1992).

[1]Department of Engineering, University of Cambridge, UK [2]Alan Turing Institute, London, UK [3]University College London, UK. Correspondence to: Tim Pearce <tp424@cam.ac.uk>.

Whilst this is an attractive solution, the number of parameters in modern NNs can be in the order of millions, rendering full Bayesian inference impractical. One ongoing line of research within NN uncertainty aims to develop faster, more practical, approximate Bayesian inference schemes for NNs (Graves, 2011; Hernández-Lobato & Adams, 2015; Blundell et al., 2015).

Other approaches to uncertainty quantification aim to achieve Bayesian-like behaviour through clever use of existing architectures, searching for techniques that have minimal impact on usage. *MC dropout* is one popular approach. The original work (Gal & Ghahramani, 2015) showed that dropout, under certain approximations and assumptions, performs Variational Inference (VI). However, it has attracted criticism and a counterexample (Osband et al., 2016; Osband, 2016), and in our own experiments failed to produce satisfactory results (figure 1).

Ensembling aggregates the estimates of multiple individual NNs, each originally having a different parameter initialisation. The variance of the ensemble's predictions may be interpreted as its uncertainty. However, justification provided for current methods is more intuitive than analytical: it is assumed each individual NN will converge to similar results where data has been observed, but predictions will be diverse elsewhere. It has a long history dating back to Heskes (1996), and remains popular today (Osband et al., 2016; Lakshminarayanan et al., 2017; Pearce et al., 2018). It has often been proposed in conjunction with boostrap resampling, however certain works have found that this harms performance (Lee et al., 2015).

The analytical justification for ensembling is that the initialising distribution can be interpreted as a prior, and a trained NN can be viewed as sampling from the posterior. Two prominent recent works have challenged this:

- Gal (2016) provides a counterexample to ensembling, arguing a case where all members of an ensemble would reach the same solution with no diversity.

- Osband (2016) noted *'it may be necessary to maintain some more rigorous notion of prior'*, suggesting the synthesis of data as one feasible approach.

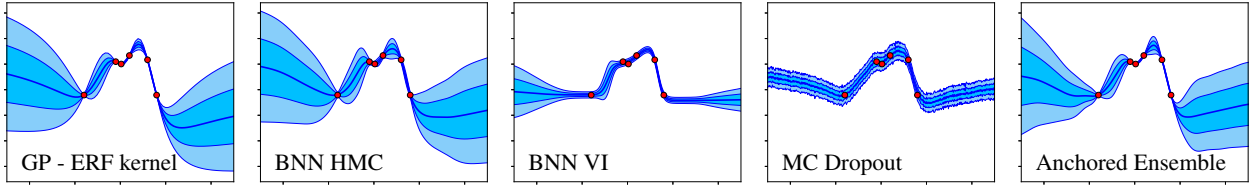

*Figure 1.* Bayesian inference in wide, single-layer NNs using various methods on six generated data points. Shades of blue show standard deviations of the predictive distributions. Whilst full inference using GP or HMC is desirable, these methods do not scale well. VI and MC dropout do not maintain correlations between parameters and appear a poor substitute. Anchored ensembles provide a pragmatic solution.

These two comments both target the concern that regardless of the initialising prior distribution, training drives a NN directly to the likelihood distribution. In the simplest cases, this can result in zero diversity amongst ensemble members.

Duvenaud & Adams (2016) investigated the use of early stopping in catching the distribution as it shifts from prior to likelihood, arguing that if the number of training steps is set correctly the distribution captured will approximate that of the posterior.

This work considers an alternative method for capturing the posterior from an ensemble of NNs that requires one minor modification to the usual implementation - **instead of regularising parameters about zero, they are regularised about their initialisation values**. In this way, the ensemble as a whole is regularised about zero, whilst the diversity of each member of the ensemble is maintained. We are aware of only one other work in which parameters were regularised around non-zero values - this was in a multi-task setting, where NNs were encouraged not to forget prelearned tasks (Kirkpatrick et al., 2015).

This minor modification has major consequences. We provide a proof (section C) that, assuming the joint likelihood of parameters is a multivariate Gaussian, and maximum a posteriori (MAP) estimates are returned for the parameters, full Bayesian Inference is performed. We also show that this multivariate Gaussian assumption holds exactly for the final layer weights of a NN. Given well-established results on the relationship between BNNs and GPs (Neal, 1996), this means that **an anchored ensemble of NNs converges to the corresponding GP as both the size of each NN, as well as the number of NNs, tend to infinity**.

Figure 1 provides a visual comparison between common Bayesian inference methods and this work's proposed method. Included are a BNN with Hamiltonian Monte Carlo (HMC) (Neal, 1997; 2012), BNN with VI, NN with MC dropout, the proposed ensembling method (henceforth termed *anchored ensemble*), and the equivalent GP using the covariance function derived by Williams (1996). We believe that analytical inference using **a GP with kernel corresponding to a NN sets an upper limit on the quality of uncertainty estimate that may be achieved by that NN**. Methods in figure 1 should therefore be assessed relative to that of the GP.

### 1.2. Uncertainty to Drive RL Exploration

Reinforcement learning (RL) involves the use of exploration mechanics to reveal the structure of an environment to an agent. In order for the agent to learn a policy that maximizes its reward, it needs to have seen a sufficient amount of the state-space. Pragmatic methods that have worked well empirically are heuristic-based, perhaps the most popular being $\epsilon$-greedy, where the agent chooses the action with highest Q-value with probability $(1 - \epsilon)$ else selects from all possible actions uniformly at random (Montague, 1999). $\epsilon$ is a hyperparameter chosen dependent on the specific environment. It is generally adapted over time, often beginning at $1.0$ and decreasing linearly over episodes.

If the agent is able to quantify how certain it is that an action is optimal, it can use this information to perform exploration in a more principled manner: choosing to explore when less certain of an action, exploiting otherwise. Thompson sampling (Thompson, 1933) provides a simple rule for this behaviour: an agent should choose an action with probability equal to the probability it believes it is the optimal action. This has been used with success (Gal & Ghahramani, 2015; Osband et al., 2016).

Uncertainty estimates have further uses in RL, providing some measure of interpretability to understand policies (section 4). We are also interested in testing its applicability in prioritised experience replay (Schaul et al., 2015), where actions the agent was less certain about could be replayed with higher priority.

## 2. The Problem with Ensembling

This section motivates the need for anchored ensembles. There is a temptation to interpret an initialised, untrained NN as a sample from a prior distribution. We hence refer to this as a *quasi-prior*. As training proceeds, the parameters drift from this initialising distribution toward the likelihood distribution. With nothing to anchor them to their prior distribution, their final distribution may end up independent of the initialising distribution.

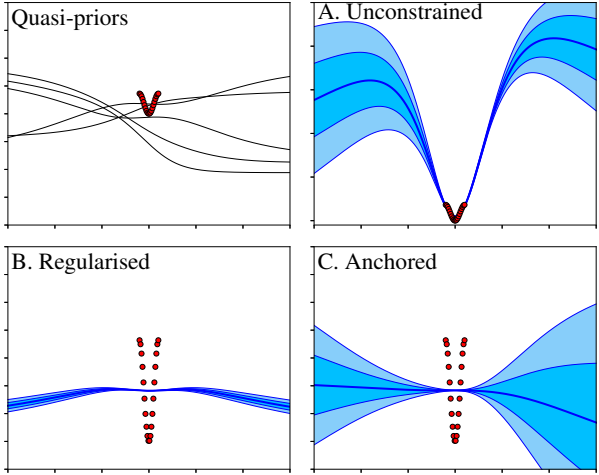

*Figure 2.* To regularise or not to regularise: regularisation reduces diversity in the ensemble (B), however removing it overfits the data (A). Anchored ensembling provides a solution (C). $x$-axis is fixed, $y$-axis has been scaled as needed.

For a single NN regularisation addresses this. Let $\boldsymbol{\theta}$ represent a flattened vector of all NN parameters, $\lambda$ a regularisation coefficient, $\hat{\mathbf{y}}$ a NN prediction and $\mathbf{y}$ the true value for $N$ data points. The regularised loss function is given by,

$$Loss = \frac{1}{N}\sum_{i=1}^{N}(\mathbf{y}_i - \hat{\mathbf{y}})^2 + \frac{\lambda}{N}||\boldsymbol{\theta}||^2. \quad (1)$$

Parameters minimising this loss can be interpereted as MAP estimates with a normal prior centered at zero (MacKay, 2005). This is a useful result for single NNs, since output functions are encouraged to be smooth. Unfortunately this benefit does not carry to ensembles: it is desirable that they contain diversity in regions of unobserved data, and not consistent smooth interpolations shared by all NNs.

This leaves a practitioner with a dilemma: don't regularise and risk overfitting, or regularise and risk too little diversity in the ensemble. Figure 2 illustrates the above discussion and demonstrates the effectiveness of anchored ensembles.

## 3. Methodology

### 3.1. Bayesian Inference with MAP Anchoring

This section introduces, in broad strokes, the mechanism behind anchored ensembling. Bayesian inference generally considers three distributions; prior, likelihood and posterior.

**Algorithm 1** Exploration in RL using anchored ensembles
**Input:** Multiple NNs with varying initialisations, $\boldsymbol{\theta}_0$, an environment with exactly 2 actions $a$, returning states $s$ and rewards $r$, a buffer $B$ storing experiences.
**for each** episode
  **while** not done
    **for each** NN **in** ensemble
      $Q\_0_i$ = get Q-value for action 0 from $\text{NN}_i(s_t)$
      $Q\_1_i$ = get Q-value for action 1 from $\text{NN}_i(s_t)$
    **end for**
    $Q\_0_\mu, Q\_0_{\sigma^2}$ = mean & variance of $Q\_0$'s
    $Q\_1_\mu, Q\_1_{\sigma^2}$ = mean & variance of $Q\_1$'s
    Fit $Q\_0$'s & $Q\_1$'s with a Gaussian distribution
    $p\_0 = Pr(\mathcal{N}(Q\_0_\mu, Q\_0_{\sigma^2}) \geq \mathcal{N}(Q\_1_\mu, Q\_1_{\sigma^2}))$
    $a_t$ = select $a_0$ with probability $p\_0$, else $a_1$
    Perform action $a_t$ and receive $s_{t+1}, r_t$
    Append $(a_t, s_t, s_{t+1}, r_t)$ to $B$
  **end while**
  Sample from $B$
  Train all NNs on same sample using eq. 2
**end for**

Here, a further distribution is introduced, which we term the *anchoring distribution*, identical in position to a hyperprior, but with a different role. Inference proceeds by sampling initial parameters, $\boldsymbol{\theta}_0$, from the anchoring distribution. MAP estimates for $\boldsymbol{\theta}$ are then returned with a prior centered about $\boldsymbol{\theta}_0$. For a NN this is done by finding parameters minimising a modified loss,

$$Loss_{anchor} = \frac{1}{N}\sum_{i=1}^{N}(\mathbf{y}_i - \hat{\mathbf{y}})^2 + \frac{\lambda}{N}||\boldsymbol{\theta} - \boldsymbol{\theta}_0||^2. \quad (2)$$

**Theorem 1.** *Assume that the joint likelihood of model parameters follows a multivariate normal distribution, that the prior is normally distributed, and that there exists a mechanism by which optimal MAP parameter estimates can be returned. The proposed anchored inference scheme provides a consistent estimator of the posterior.*

This inference procedure is not specific to NNs, and we are exploring use in other contexts. Proof of theorem 1 is given in appendix C, where we also show that the distributional assumption holds exactly for the final weight layer of a NN. Figure 8 illustrates the accuracy of the procedure.

### 3.2. Thompson Sampling with Ensembles

Algorithm 1 summarises how Thompson sampling is performed with an anchored ensemble of NNs. Note that this differs slightly from the scheme used in Osband & Adams (2016), where a single NN was used for each episode.

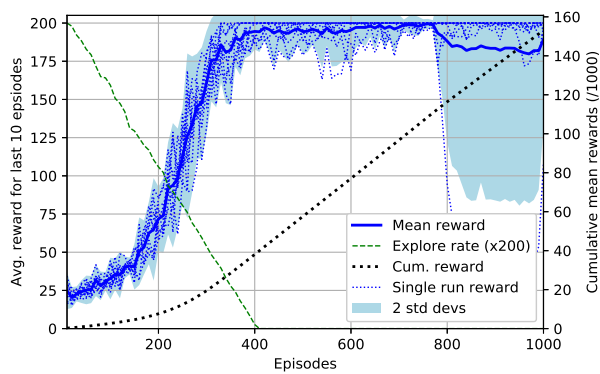

*Figure 3.* Single $\epsilon$-greedy NN, with $\epsilon$ annealed linearly over 400 episodes. Note the sigmoidal shape, and instability at episode 800.

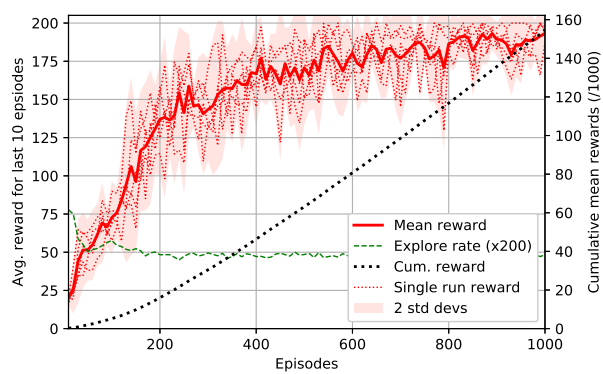

*Figure 4.* An anchored ensemble of ten NNs, exploring via Thompson sampling. Note the decayed exponential shape.

## 4. Experiments in Reinforcement Learning

As an initial case study, experiments[1] were run on the *Cart Pole* control problem[2], where an agent must learn to move a trolley right or left in order to keep a pole from falling. A reward of 1 is received for every time step the pole is kept aloft, with episodes running for a maximum of 200 time steps. Q-learning with single layer NNs was employed.

$\epsilon$-greedy was compared with Thompson sampling, which utilised uncertainty estimates generated by an anchored ensemble. Results are in figures 3 & 4, which show average reward, cumulative reward, and the number of exploration activities performed (normalised to 200). Results were averaged over five individual training runs.

Both methods gained similar cumulative reward over 1,000 episodes. The obvious difference between the two methods is the shape of the learning curve: $\epsilon$-greedy has worse initial performance as it randomly tries out bad moves, but quickly learns a successful, although basic strategy - it favours one particular direction regardless of the initial state, and gradually accelerates in that direction. This gives the learning curve a sigmoidal shape. Conversely, Thompson sampling produces an exponential decay shape, with learning beginning immediately, but not experiencing the same sudden improvement. Generally Thompson sampling appears to promote more stable ($\epsilon$-greedy becomes unstable around episode 800), and steady learning behaviour, also discovering a more a sensible policy.

We trialled several experiment variants - see appendix B for plots. Significantly, an anchored ensemble did not improve performance relative to a regular ensemble - we note that anchoring promotes Bayesian behaviour which prevents overfitting, but this may not be an advantage in noiseless, deterministic environments. The number of NNs in the ensemble also had little impact. We suspect that given a more complex problem, these may become increasingly important, and also that a higher performance ceiling could be realised as found in Osband & Adams (2016).

To assess the agent's learnt confidence in actions, we considered three categories of state as in appendix A: *non-critical* (pole is balanced with no apparent optimal action), *critical* (pole is tilted so that a wrong action quickly leads to failure), and *rare* (a state unlikely to ever be observed). The distributions over Q-values for each action show that the agent explores in the non-critical case (both action probabilities are around $0.5$), and the agent takes correct action with high probability in the critical case. In the rare case, the action probabilities are again around $0.5$ but here the variance of each action distribution remains high, showing the agent is uncertain of either action. This demonstrates that the agent develops a policy to explore in scenarios which, from experience, are not critical to obtaining rewards, allowing safe exploration whilst exploiting existing knowledge.

## 5. Conclusions and Ongoing Work

This paper proposed one minor modification to ensembling methodology that regularises individual NNs around their initialised parameter values rather than zero. This allows a notion of the prior distribution to be maintained, addressing a major criticism of ensembles, and opening a cheap route to Bayesian inference in cases where MAP parameter estimates are easily returned.

Small-scale experiments in RL showed that anchored ensembles provide a pragmatic approach to Thompson sampling, which provided several benefits over $\epsilon$-greedy.

This is ongoing work, and we are actively pursuing ideas raised in this working paper. These include developing broader proofs, and analysing how quickly anchored ensembles converge to their corresponding GP. We are assessing their applicability in deep, complex NNs, both in supervised learning tasks and in more challenging RL environments.

---

[1] https://github.com/TeaPearce }
[2] OpenAI Gym - https://gym.openai.com

## Acknowledgements

The authors thank EPSRC for funding (EP/N509620/1), the Alan Turing Institute for accommodating the lead author during his work (TU/D/000016), and Microsoft for Azure credits.

## A. Confidence in Actions Analysis

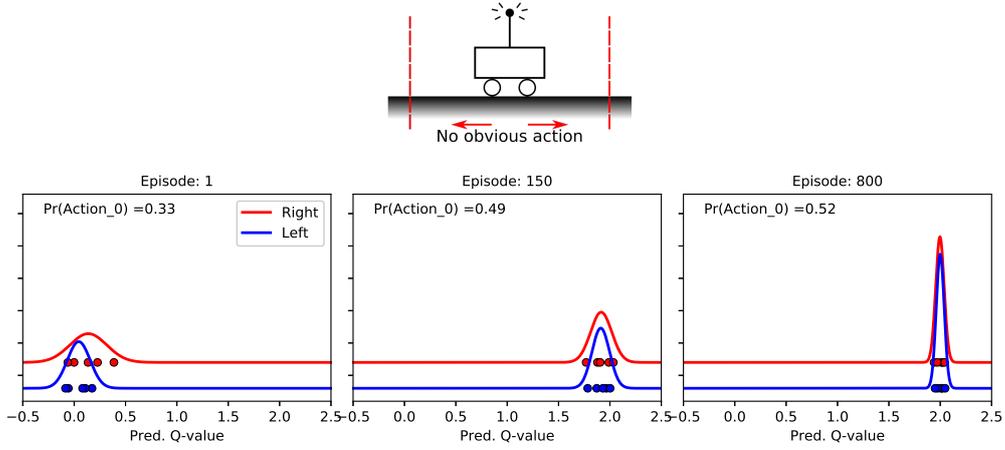

Figure 5. The agent's confidence in actions for **non-critical** states. The agent learns that there is no clear optimal action, and it is confident of this (low variance of each distribution).

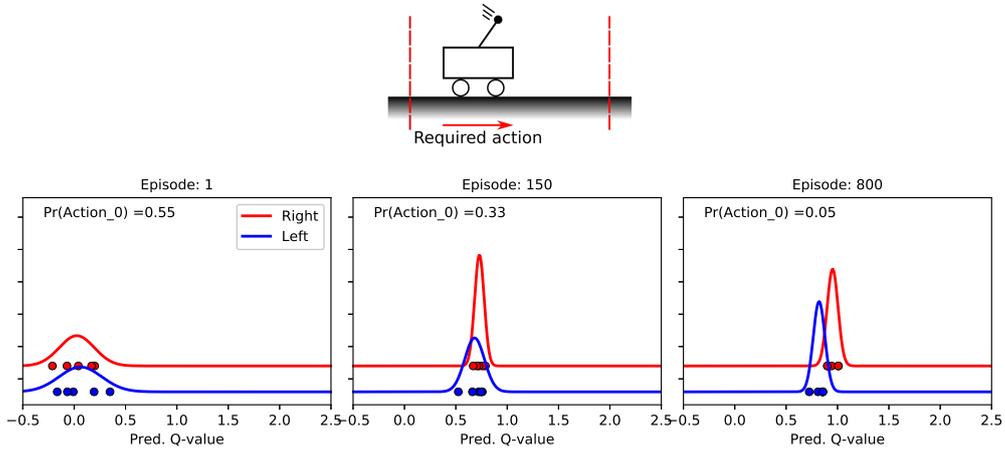

Figure 6. For **critical** states the agent learns that action qualities are significantly different, and with reasonable confidence.

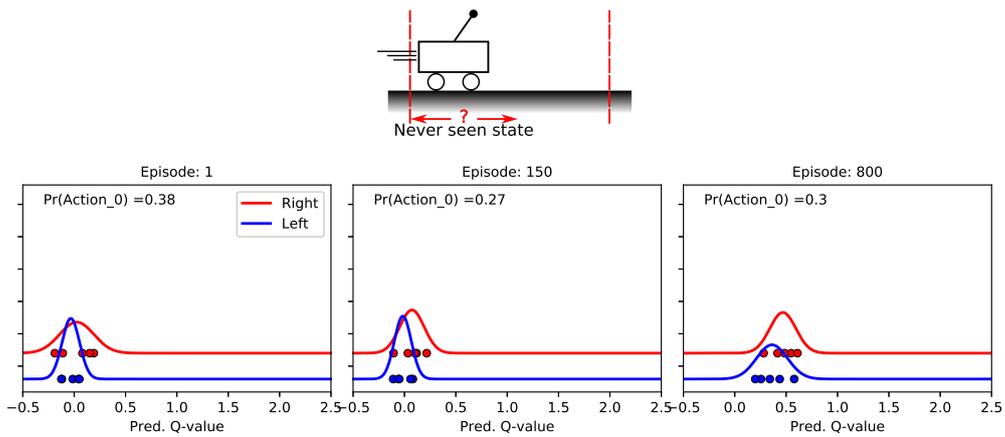

Figure 7. For **rare** states the distributions remain broad and probabilities of selection approximately equal.

## B. Additional Plots

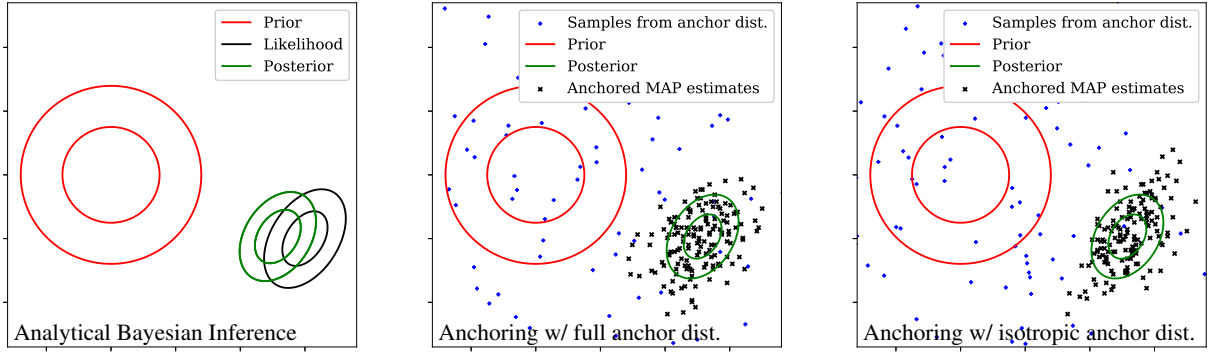

*Figure 8.* Performing Bayesian inference with anchored MAP estimates. 150 samples are shown for the case where the exact anchoring distribution is used (middle), as well as for an approximate isotropic anchoring distribution (right).

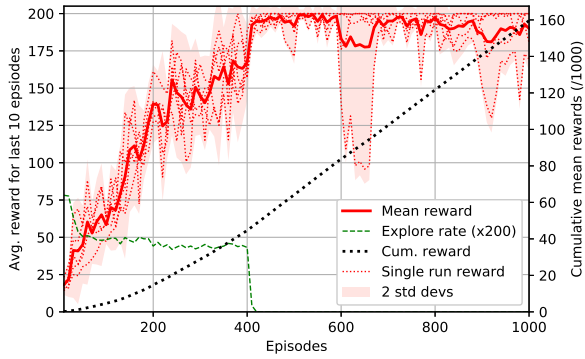

*Figure 9.* Anchored ensemble of ten NNs, Thompson sampling up to episode 400, fully greedy behaviour afterwards.

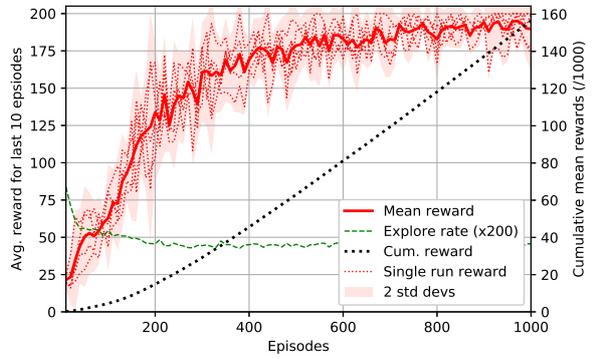

*Figure 10.* Unanchored ensemble of ten NNs, Thompson sampling.

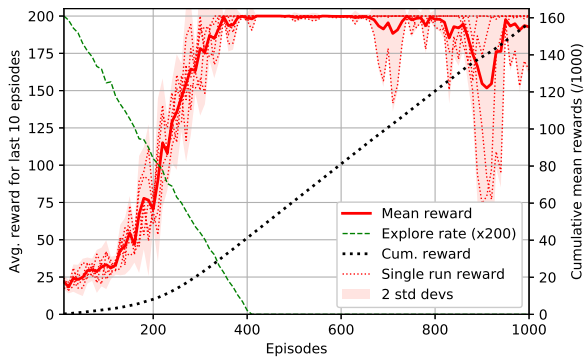

*Figure 11.* Anchored ensemble of ten NNs, $\epsilon$-greedy exploration with $\epsilon$ annealed as in figure 3.

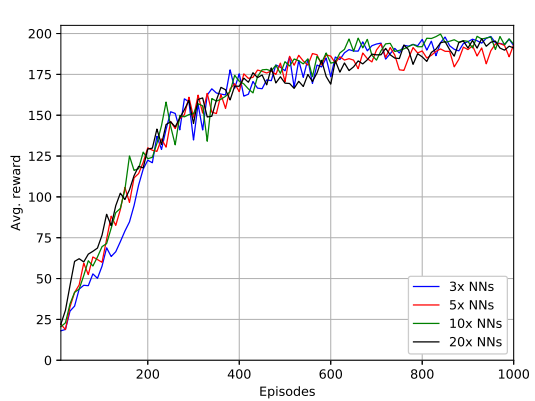

*Figure 12.* Comparison of anchored ensembles of differing sizes.

# C. Proofs

**Theorem 1.** *Assume that the joint likelihood of model parameters follows a multivariate normal distribution, that the prior is normally distributed, and that there exists a mechanism by which optimal MAP parameter estimates can be returned. The proposed anchored inference scheme provides a consistent estimator of the posterior.*

*Proof.* Consider prior and (normalised) likelihood distributions, both multivariate normal,

$$P(\theta) = \mathcal{N}(\boldsymbol{\mu}_{prior}, \boldsymbol{\Sigma}_{prior}), \tag{3}$$

$$\frac{P(\mathcal{D}|\theta)}{P(\mathcal{D})} = \mathcal{N}(\boldsymbol{\mu}_{like}, \boldsymbol{\Sigma}_{like}), \tag{4}$$

with posterior calculated by Bayes rule,

$$P(\theta|\mathcal{D}) = \frac{P(\mathcal{D}|\theta)P(\theta)}{P(\mathcal{D})}. \tag{5}$$

**Standard Result 1:** (§8.1.8, The Matrix Cookbook, 2008) If both prior and likelihood are multivariate normal, the posterior is also normal and available in closed form,

$$P(\theta|\mathcal{D}) = \mathcal{N}(\boldsymbol{\mu}_{post}, \boldsymbol{\Sigma}_{post}), \tag{6}$$

$$\boldsymbol{\Sigma}_{post} = (\boldsymbol{\Sigma}_{prior}^{-1} + \boldsymbol{\Sigma}_{like}^{-1})^{-1}, \tag{7}$$

$$\boldsymbol{\mu}_{post} = \boldsymbol{\Sigma}_{post}\boldsymbol{\Sigma}_{prior}^{-1}\boldsymbol{\mu}_{prior} + \boldsymbol{\Sigma}_{post}\boldsymbol{\Sigma}_{like}^{-1}\boldsymbol{\mu}_{like}. \tag{8}$$

We now introduce an anchoring distribution which we enforce as multivariate normal,

$$\theta_{anc} \sim P(\theta_{anc}) = \mathcal{N}(\boldsymbol{\mu}_{anc}, \boldsymbol{\Sigma}_{anc}). \tag{9}$$

This is used as described in the text so that samples are taken from the anchoring distribution, with a prior centered at each,

$$P(\theta) = \mathcal{N}(\theta_{anc}, \boldsymbol{\Sigma}_{prior}), \tag{10}$$

where $\boldsymbol{\Sigma}_{prior}$ is unchanged from eq. 3.

Denote $\theta'$ as the MAP estimates given this prior and the original likelihood from eq. 4.

We must show three things regarding $\theta'$:

- that its distribution follows a multivariate normal,

$$\theta' \sim P(\theta') = \mathcal{N}(\boldsymbol{\mu}', \boldsymbol{\Sigma}'), \tag{11}$$

- that $\boldsymbol{\mu}_{anc}$ & $\boldsymbol{\Sigma}_{anc}$ can be selected in such a way that the mean of the distribution is equal to that of the original posterior

$$\boldsymbol{\mu}' = \boldsymbol{\mu}_{post}, \tag{12}$$

- and also so that the covariance of the distribution is equal to that of the original posterior

$$\boldsymbol{\Sigma}' = \boldsymbol{\Sigma}_{post}. \tag{13}$$

We make use of the following standard result.

**Standard Result 2:** (§8.1.4, The Matrix Cookbook, 2008) For a random variable, $\mathbf{x}$, normally distributed and with an affine transformation applied,

$$\mathbf{x} \sim \mathcal{N}(\boldsymbol{\mu}_c, \boldsymbol{\Sigma}_c), \tag{14}$$

$$\mathbf{y} = \mathbf{A}\mathbf{x} + \mathbf{b}, \tag{15}$$

$\mathbf{y}$ will also be normally distributed with parameters as follows,

$$\mathbf{y} \sim \mathcal{N}(\mathbf{A}\boldsymbol{\mu}_c + \mathbf{b}, \mathbf{A}\boldsymbol{\Sigma}_c\mathbf{A}^T). \tag{16}$$

Consider a single sample from the anchoring distribution, $\theta^*_{anc}$, that is adopted by the prior as,

$$\mathcal{N}(\theta^*_{anc}, \boldsymbol{\Sigma}_{prior}). \tag{17}$$

Denote $\theta^*_{post}$ as the MAP parameter estimate of the posterior formed by this prior and the likelihood in eq. 4. We have already seen that the posterior is also normally distributed, and its mean, which is also the MAP estimate, is given by combining eq. 7, 8 & 17,

$$\theta^*_{post} = (\boldsymbol{\Sigma}_{prior}^{-1} + \boldsymbol{\Sigma}_{like}^{-1})^{-1}\boldsymbol{\Sigma}_{prior}^{-1}\theta^*_{anc} + \\ (\boldsymbol{\Sigma}_{prior}^{-1} + \boldsymbol{\Sigma}_{like}^{-1})^{-1}\boldsymbol{\Sigma}_{like}^{-1}\boldsymbol{\mu}_{like}. \tag{18}$$

Defining for convenience,

$$\mathbf{A}_1 = (\boldsymbol{\Sigma}_{prior}^{-1} + \boldsymbol{\Sigma}_{like}^{-1})^{-1}\boldsymbol{\Sigma}_{prior}^{-1}, \tag{19}$$

$$\mathbf{b}_1 = (\boldsymbol{\Sigma}_{prior}^{-1} + \boldsymbol{\Sigma}_{like}^{-1})^{-1}\boldsymbol{\Sigma}_{like}^{-1}\boldsymbol{\mu}_{like}, \tag{20}$$

this becomes,

$$\theta^*_{post} = \mathbf{A}_1\theta^*_{anc} + \mathbf{b}_1, \tag{21}$$

which is the same form as eq. 15. Hence, from Standard Result 2, if $\theta^*_{anc}$ is normally distributed, $\theta^*_{post}$ **will also be normally distributed**.

Regarding the mean of $\theta^*_{post}$, we have,

$$\mathbb{E}[\theta^*_{post}] = \mathbb{E}[\mathbf{A}_1 \theta^*_{anc} + \mathbf{b}_1] \tag{22}$$

$$= \mathbf{A}_1 \mathbb{E}[\theta^*_{anc}] + \mathbf{b}_1. \tag{23}$$

By choosing the anchoring distribution to be centered about the original prior,

$$\mathbb{E}[\theta^*_{anc}] = \mathbb{E}[\theta_{prior}] = \boldsymbol{\mu}_{prior}, \tag{24}$$

we have,

$$\mathbb{E}[\theta^*_{post}] = \boldsymbol{\Sigma}_{post}\boldsymbol{\Sigma}^{-1}_{prior}\boldsymbol{\mu}_{prior} + \boldsymbol{\Sigma}_{post}\boldsymbol{\Sigma}^{-1}_{like}\boldsymbol{\mu}_{like}, \tag{25}$$

which is consistent with eq. 8 and proves that **the means of the distributions are aligned**.

Finally we consider the variance of $\theta^*_{post}$, which we wish to equal $\boldsymbol{\Sigma}_{post}$ by choosing $\boldsymbol{\Sigma}_{anc}$. Using the form from eq. 21 we find,

$$\mathbb{V}ar[\theta^*_{post}] = \mathbb{V}ar[\mathbf{A}_1 \theta^*_{anc} + \mathbf{b}_1] \tag{26}$$

$$= \mathbf{A}_1 \mathbb{V}ar[\theta^*_{anc}]\mathbf{A}^T_1 = \mathbf{A}_1 \boldsymbol{\Sigma}_{anc} \mathbf{A}^T_1 \tag{27}$$

We require the following result,

$$\boldsymbol{\Sigma}_{post} = (\boldsymbol{\Sigma}^{-1}_{prior} + \boldsymbol{\Sigma}^{-1}_{like})^{-1} = \mathbf{A}_1 \boldsymbol{\Sigma}_{anc} \mathbf{A}^T_1. \tag{28}$$

Denoting, $\mathbf{C} := \boldsymbol{\Sigma}^{-1}_{prior} + \boldsymbol{\Sigma}^{-1}_{like}$, and rearranging,

$$\boldsymbol{\Sigma}_{anc} = \mathbf{A_1}^{-1}\mathbf{C_1}^{-1}\mathbf{A_1^T}^{-1} \tag{29}$$

$$= \mathbf{C}\boldsymbol{\Sigma}^{-1}_{prior}\mathbf{C}^{-1}\boldsymbol{\Sigma}^T_{prior}\mathbf{C^T}^{-1}. \tag{30}$$

If $\boldsymbol{\Sigma}_{prior}$ is selected to be diagonal and isotropic, it can be replaced by $\lambda_1 \cdot \mathbb{I}$, which reduces to,

$$\boldsymbol{\Sigma}_{anc} = \lambda^3_1 \mathbb{I} + \lambda^2_1 \boldsymbol{\Sigma}^{-1\mathbf{T}}_{like}. \tag{31}$$

$\boldsymbol{\Sigma}_{like}$ may contain non-diagonal entries which means $\boldsymbol{\Sigma}_{anc}$ is not necessarily diagonal. However could be approximated by an isotropic distribution,

$$\boldsymbol{\Sigma}_{like} \approx \lambda_2 \mathbb{I} \tag{32}$$

$$\boldsymbol{\Sigma}_{anc} \approx (\lambda^3_1 + \lambda^2_1 \lambda_2)\mathbb{I}. \tag{33}$$

With proper selection of $\lambda_1$ and $\lambda_2$, **the variance of the distributions will be aligned**.

□

Figure 8 illustrates the sampling procedure for 150 samples using both the full anchored covariance matrix in eq. 31, and the isotropic approximation in eq. 33, for a two dimensional toy example.

Practically speaking, a straightforward approach to setting $\boldsymbol{\Sigma}_{anc}$ is to consider a single variable, $\lambda_3 = \lambda^3_1 + \lambda^2_1 \lambda_2$, which can be found via cross validation as for other hyperparameters.

**Proposition 1.** *The parameters of the final weight layer of a NN have multivariate normal likelihood.*

*Proof.* This can be seen relatively simply using the result that the coefficients of a linear model are jointly normal if a Gaussian noise model is assumed. A NN could be interpreted as providing a set of covariates which are linearly transformed by the final layer, hence these weights also have a multivariate normal likelihood. □